\documentclass[sigconf]{acmart}

\copyrightyear{2021}
\acmYear{2021}
\setcopyright{acmlicensed}
\acmConference[AI-ML Systems '21] {Doctoral Symposium at the First International Conference on AI-ML Systems}{October 21--23, 2021}{Bangalore, India.}
\acmBooktitle{Proceedings of the First International Conference on AI-ML Systems, October 21--23, 2021, Bangalore, India}
\acmPrice{15.00}
\acmISBN{978-1-4503-8446-9/21/11}

\usepackage{balance}
\usepackage{amsmath,amsfonts}
\usepackage{euler}
\usepackage{algorithmic}
\usepackage{graphicx}
\usepackage{textcomp}
\usepackage{xcolor}
\usepackage{xcolor}
\usepackage{xspace}
\usepackage[ruled, vlined, linesnumbered]{algorithm2e}
\usepackage{algorithmic}
\usepackage{graphicx}
\usepackage{amsmath}
\usepackage{amsfonts}
\usepackage{subcaption}
\usepackage{amsthm}
\usepackage{mathrsfs}
\usepackage{booktabs}
\usepackage{color}
\usepackage{hyperref}
\usepackage{lipsum}

\newcommand{\cm}[1]{\mathcal{#1}}

\newcommand{\bs}[1]{\boldsymbol{#1}}
\newcommand{\eg}{\emph{e.g.}}
\newcommand{\ie}{\emph{i.e.}}
\newcommand{\etc}{\emph{etc.}}

\newcommand{\xhdr}[1]{\vspace{0mm}\noindent{{\bf #1.}}}

\renewcommand{\comment}[1]{}
\usepackage{xcolor}
\usepackage{multirow,amsmath,booktabs,dcolumn}
\usepackage{todonotes}
\usepackage{colortbl}
\usepackage{paralist}
\usepackage{graphicx}
\usepackage{caption}

\usepackage{mathtools}

\SetAlFnt{\footnotesize}
\usepackage{xcolor,colortbl}
\newcolumntype{a}{>{\columncolor{blue!5}}c}
\newcolumntype{b}{>{\columncolor{red!5}}c}

\begin{document}
\title{Learning Neural Models for Continuous-Time Sequences}
\author{Vinayak Gupta}
\affiliation{\institution{IIT Delhi}}
\email{vinayak.gupta@cse.iitd.ac.in}

\begin{abstract}
The large volumes of data generated by human activities such as online purchases, health records, spatial mobility \etc\ are stored as a sequence of events over a continuous time. Learning deep learning methods over such sequences is a non-trivial task as it involves modeling the ever-increasing event timestamps, inter-event time gaps, event types, and the influences between events -- within and across different sequences. This situation is further exacerbated by the constraints associated with data collection \eg\ limited data, incomplete sequences, privacy restrictions \etc\ With the research direction described in this work, we aim to study the properties of continuous-time event sequences (CTES) and design robust yet scalable neural network-based models to overcome the aforementioned problems. In this work, we model the underlying generative distribution of events using marked temporal point processes (MTPP) to address a wide range of real-world problems. Moreover, we highlight the efficacy of the proposed approaches over the state-of-the-art baselines and later report the ongoing research problems.
\end{abstract}

\maketitle

\section{Motivation} \label{sec:motivation}
Continuous-time event sequences (CTES) have become pervasive across many applications ranging from healthcare~\cite{rizoiu2}, traffic~\cite{thp, imtpp}, social networks~\cite{rmtpp, coevolve}, finance~\cite{intfree, hawkes}, and sensor networks~\cite{tulone2006paq, lbrr}. Unlike images and text, the data quality of CTES is highly susceptible to the collection process \ie\ a few missing events or shorter sequences can significantly hamper the data quality and consequently the performance of the neural models trained on such datasets. This situation is further exacerbated by privacy restrictions such as GDPR~\cite{gdpr} and thus, overcoming the drawbacks in dataset quality is a non-trivial task unaddressed in the past literature.

\section{Preliminaries: MTPP}
Marked temporal point processes (MTPP) are stochastic processes realized by a sequence of asynchronous events in continuous-time. We represent an MTPP $\cm{S}_k=\{e_i=(m_i,t_i) | i \in[k] , t_i<t_{i+1}\}$, where $t_i\in\mathbb{R}^+$ is the occurrence time and $m_i\in \cm{C}$ is the discrete mark associated with the $i$-th event, with $\cm{C}$ as the set of all discrete marks. Here, $\cm{S}_k$ denotes the sequence of first $k$ events and we represent the inter-arrival times as, $\Delta_{t,k} = t_{k}-t_{k-1}$. If the sequences have a spatial component \ie\ $e_i=(m_i,t_i, d_i)$ then we represent the inter-event spatial differences as $\Delta_{d,k} = d_{k}-d_{k-1}$.

\section{Research Directions} \label{sec:research}
In this section, we highlight the CTES data-related problems addressed in our research plan and the proposed solutions. We classify them into the following categories:
\subsection{Incomplete Sequences} \label{sec:incomp}
Traditional models and inference methods for CTES assume a complete observation scenario \ie\ the events in a sequence are completely observed with no missing events -- an ideal setting and rarely applicable in real-world applications. Undeniably, any data collection procedure may not capture some events due to crawling and privacy restrictions by online platforms. As mentioned in Section~\ref{sec:motivation}, these missing events can drastically affect the quality of data and hence, the prediction performance of the learned model.

\xhdr{Proposed Solution}
In~\citet{imtpp}, we present an unsupervised model and inference method for learning neural MTPP over CTES with missing events. Specifically, we design a \textit{coupled}-MTPP approach that first models the generative processes of both -- observed events and missing events -- where the missing events are represented as latent-random variables. Later, we jointly learn the distribution of all events via MTPP using variational inference. 

The proposed model IMTPP (\textbf{I}ntermittently-observed \textbf{M}arked \textbf{T}emporal \textbf{P}oint \textbf{P}rocesses) models the generative distribution of observed events ($e_k$) and missing events ($\epsilon_r$) using MTPP denoted as $p(\bullet)$ and $q(\bullet)$ respectively. For generating future observed events, we follow a standard sampling procedure~\cite{intfree}, however, the generation process of missing events is conditioned on the history as well as the \textit{next} observed event. Specifically, we sample missing events between two observed events, till we reach the future observed event \ie\ the MTPP are -- $p(e_k)$ and $q(\epsilon_r|e_{k+1})$ respectively. We learn the parameter by maximizing a variational lower bound or evidence lower bound (ELBO) of the log-likelihood.
\begin{equation}
\mathbb{E}_q \sum_{k=0}^{K-1}\log p(e_{k+1}) - \sum_{k=0}^{K-1} KL \bigg[ q(\epsilon_r | e_{k+1}) || p_r(\epsilon_r))\bigg],
\end{equation}
where $p_r, KL$ denote prior MTPP and KL-divergence respectively.

\xhdr{Results}
We evaluated IMTPP across five datasets from different domains ranging from -- Amazon Movies, Amazon Toys, Twitter, Foursquare, and Stack-Overflow. To summarize the results, our observations were: \begin{inparaenum} \item for predicting the mark and time of events in the test set, IMTPP outperformed the state-of-the-art approaches~\cite{rmtpp, thp, intfree}, by up to 8\% across all datasets, \item for forecasting future events in a sequence, the gains IMTPP had over other baselines we consistent even for farther predictions, and \item a scalability analysis over datasets with millions of events showed that other CTES models took up to 24 hours training times, whereas the times for IMTPP were under 5 hours.\end{inparaenum}

\xhdr{Ongoing Work and Challenges}
As an extension to IMTPP, we addressed the problem of imputing missing events in a sequence. Specifically, we sampled missing events using the posterior MTPP and evaluated over synthetically deleted events in a sequence. The results, including qualitative analysis, showed that the new approach outperforms other baselines and opens up new applications of neural models and CTES. This extension is under review in a reputed journal. A major challenge with imputations is to estimate the total number of missing events between each observed event. 

\subsection{Limited Training Data}
The problem of limited training data is ubiquitous in all real-world applications ranging from recommender systems~\cite{manasi}, vision~\cite{image}, spatial models~\cite{www}  \etc\ In contrast to incomplete data, defined in Section~\ref{sec:incomp}, we regard \textit{limited} data as the problem of data-scarcity \ie\ the available data is assumed to be complete but the volume is insufficient to effectively train a deep neural network. 

\xhdr{Proposed Solution}
Transfer learning (TL) has long been proposed as a feasible solution to overcome limited data problems~\cite{bengio_tran}. Accordingly, in~\citet{reformd}, we present a transfer approach for training neural MTPP on a data-rich dataset and fine-tune the model parameters on a data-scarce dataset. In detail, we consider the problem of mobility prediction, wherein we have spatial trajectories of users across different regions and regions with data-rich mobility sequence as a \textit{source} region and the \textit{target} region with scarce mobility data.

The proposed model REFORMD(\textbf{Re}usable \textbf{F}lows f\textbf{or} \textbf{M}obility \textbf{D}ata), learns the spatial and temporal distribution of a user trajectory using normalizing flows(NFs)~\cite{shakir}. We make the trained NFs \emph{invariant} to a region, by restricting our model to learn the distribution of inter-event time intervals, $\Delta_{t,\bullet}$, and spatial distances, $\Delta_{d,\bullet}$. Since these features are independent of the underlying region, the trained NFs can be easily fine-tuned for any mobility sequence. Moreover, we use a \textit{log-normal} flow to model both distributions \eg\ for time, the generative distribution for future events is learned as:
\begin{equation}
p_t(\Delta_{t, k+1} | \bs{s}_k) = \texttt{LogNormal} \big(\mu_t(\bs{s}_k), \sigma^2_t(\bs{s}_k)\big),
\end{equation}
with $[\mu_t(\bs{s}_k), \sigma^2_t(\bs{s}_k)] = [\bs{W}_{1} \bs{s}_k + \bs{b}_1, \bs{W}_{2} \bs{s}_k + \bs{b}_2]$ denote the \textit{mean} and \textit{variance} of the time distribution. $\bs{s}_k$ is the output of neural MTPP, $\bs{W}_{\bullet}$ and $\bs{b}_{\bullet}$ are trainable parameters. We sample the time of future events as $\Delta_{t, k+1} \thicksim p_t$ and a similar procedure is followed for spatial flows. All model parameters are trained by maximizing the location recommendation accuracy and the likelihood of time and distance of events, Moreover, we use a standard procedure to fine-tune the model parameters on the sequences from the target region.

\xhdr{Results}
We evaluated our model across eight mobility datasets from the US and Japan and our observations were: \begin{inparaenum} \item for location recommendation and event-time prediction, REFORMD outperformed other MTPP models by up to 20\% and 23\% respectively \item It also demonstrated better and faster convergence on target dataset than other approaches, and \item even in the absence of spatial flows, we outperform other methods by 3\% for item recommendation and 14\% for time prediction across four datasets from Amazon. \end{inparaenum}

\xhdr{Ongoing Work and Challenges}
A crucial drawback of our proposed model is its standard transfer learning procedure, whereas, modern neural models deploy a meta-learning~\cite{maml} procedure to transfer model parameters. Therefore as future work, we plan to combine meta-learning with NF to design robust MTPP models.

\vspace{-0.1cm}
\subsection{Latent Features}
The latent features that play a crucial role in CTES are inter-event and inter-sequence relationships. \citet{coevolve} show that these relationships imitate an information diffusion process \eg\ social preferences, community formations \etc\ and can be captured using MTPP models. Accordingly, we address the problem of community detection in the absence of the social network of a user. Specifically, we assign communities to users in a network based on their mobility preferences. 

\xhdr{Proposed Solution}
In our work~\cite{ank} with Ankita, a former Ph.D. student, we devise a spatial-temporal point process method to learn the diffusion process to assign communities to users in the network. In detail, we train a self-exciting MTPP for each user with a common excitation matrix. Later, we assign a community to a user based on their personal MTPP and the shared matrix. We learn the latent community of users and our model parameters using stochastic variational inference.

\xhdr{Results and Challenges}
The results across two spatial mobility datasets show that our model achieves improvements of up to 27\% in location prediction and 8\% for community prediction in comparison to other neural models. Though we use certain heuristics to estimate the user community, evaluating it in the absence of true labels is still an open problem.

\section{Ongoing Work} \label{sec:ongoing}
Here we highlight the ongoing works that are under review.

\xhdr{Time Series as Graphs} In this work, we represent a time series as a graph with weighted temporal edges. Later, we learn the dynamics of sequences using graph neural networks. Moreover, we overcome data scarcity using a meta-learning algorithm. 

\xhdr{Retrieving CTES} Due to the disparate nature of CTES with events containing marks and if applicable, spatial features, the problem of sequences retrieval has been left undressed by the past literature. In this work, we propose a self-attention MTPP model for retrieving a similar sequence from a corpus, given a query sequence.

\xhdr{Time Series and Databases} In~\cite{arora2021bert} with Garima, a former Ph.D. student, we learn the embedding of entities in temporally evolving relational databases. 

\vspace{-0.1cm}
\section{Conclusion}
In this paper, we present the research directions, possible data-related problems, and solutions for learning neural models on continuous time sequences. In addition, we highlighted the ongoing works that are under review. As possible progress till October, we plan to get better results for meta-learning-based training of MTPP and notifications for the papers under review.

\clearpage
\bibliographystyle{ACM-Reference-Format}
\balance
\bibliography{main}

\end{document}